\documentclass[10pt,twocolumn,letterpaper]{article}

\usepackage{wacv}              

\usepackage{graphicx, amsmath, amssymb, booktabs, algorithm, algpseudocode, comment, caption, breqn, stmaryrd, multirow, wrapfig}

\usepackage[pagebackref,breaklinks,colorlinks]{hyperref}

\usepackage{ifthen}
\newboolean{WithProof}
\setboolean{WithProof}{true}
\newcommand{\ProofLocation}{
\ifthenelse{\boolean{WithProof}}{
Appendix
}{
\url{https://arxiv.org/abs/2210.10605}}
}

\usepackage[capitalize]{cleveref}
\crefname{section}{Sec.}{Secs.}
\Crefname{section}{Section}{Sections}
\Crefname{table}{Table}{Tables}
\crefname{table}{Tab.}{Tabs.}

\usepackage{xcolor}

\newcommand{\mycomment}[1]{\textcolor{gray}{// #1}}

\def\wacvPaperID{1671} 
\def\confName{WACV}
\def\confYear{2024}

\usepackage{pifont}
\newcommand{\xmark}{\ding{55}}%

\newcommand{\xb}{\ensuremath{\boldsymbol{x}}}
\newcommand{\xZeroHat}{\ensuremath{\widehat{x}_0(t)}}

\begin{document}

\title{Fast Diffusion EM: a diffusion model for blind inverse problems with application to deconvolution}

\author{Charles Laroche\\
GoPro \& MAP5 \\
{\tt\small charles.laroche@u-paris.fr}
\and
Andrés Almansa\\
CNRS \& Université Paris Cité\\
{\tt\small andres.almansa@parisdescartes.fr}
\and
Eva Coupete\\
GoPro\\
{\tt\small ecoupete@gopro.com}
}

\maketitle

\begin{abstract}
Using diffusion models to solve inverse problems is a growing field of research. Current methods assume the degradation to be known and provide impressive results in terms of restoration quality and diversity. In this work, we leverage the efficiency of those models to jointly estimate the restored image and unknown parameters of the degradation model such as blur kernel. In particular, we designed an algorithm based on the well-known Expectation-Minimization (EM) estimation method and diffusion models. Our method alternates between approximating the expected log-likelihood of the inverse problem using samples drawn from a diffusion model and a maximization step to estimate unknown model parameters. For the maximization step, we also introduce a novel blur kernel regularization based on a Plug \& Play denoiser. Diffusion models are long to run, thus we provide a fast version of our algorithm. Extensive experiments on blind image deblurring demonstrate the effectiveness of our method when compared to other state-of-the-art approaches.
Our code is available at \href{https://github.com/claroche-r/FastDiffusionEM}{https://github.com/claroche-r/FastDiffusionEM}.
\end{abstract}

\vspace{-0.5cm}
\section{Introduction}
Image restoration aims to recover information that has been obscured by various degradations such as blur, noise, or compression artifacts. 
Deep-learning-based methods have revolutionized the field of image restoration by achieving impressive results in various tasks. They leverage the power of deep neural network architectures to learn a mapping between training data~\cite{dong_learning_2014, zhang_beyond_2017, zhang_learning_2018}. This data-driven approach allows deep-learning models to capture intricate patterns and relationships within the image data, enabling them to restore images with superior quality and perceptual fidelity~\cite{zamir_multi-stage_2021, liang_swinir_2021}. 
On the other hand, model-based approaches express the image restoration problem as an inverse problem and exploit the degradation process structure to design regularizations and optimization algorithms to find the optimal reconstruction~\cite{perrone_total_2014}. They usually offer more control, flexibility, and interpretability. However, model-based approaches highly rely on the knowledge of the degradation forward process limiting their usefulness in practical applications. 
Some strategies try to bring the best of both worlds such as Plug-and-Play methods or deep unfolding networks~\cite{ryu_plug-and-play_2019, zhang_deep_2020, hurault2022, laroche_deep_2023, kamilov2023}.
One of the challenges behind inverse problems comes from their ill-posedness. In fact, for a single degraded image, there generally exist multiple plausible solutions. A common approach is to generate a single restored image that minimizes the mean squared error, but it does not allow the models to generate or hallucinate high-quality details~\cite{whang_deblurring_2022, saharia_image_2023}. There is a growing interest in the field of image restoration to design models that can generate all the space of plausible solutions. Those models include Generative Adversarial Networks~\cite{goodfellow_generative_2014, mirza2014conditional}
, conditional or PnP Diffusion Models~\cite{pmlr-v37-sohl-dickstein15, saharia_image_2023, kawar_denoising_2022} or Langevin dynamics~\cite{laumont2022langevin}. 
\begin{figure}[b]
    \vspace{-15pt}
    \centering
    \includegraphics[width=0.7\linewidth]{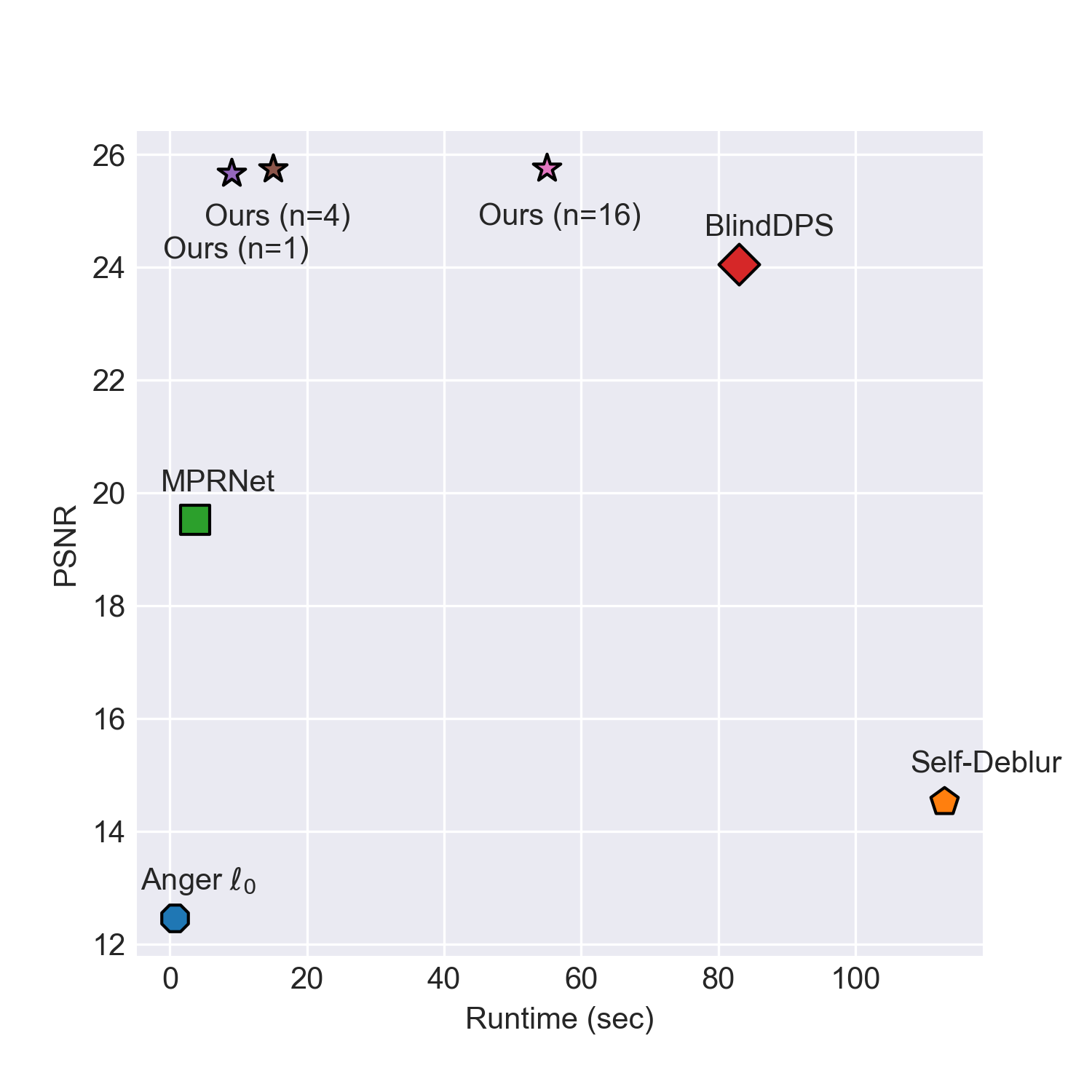}
    \caption{Performance comparison of the different models using the PSNR metric depending on the runtime, ``Ours'' corresponds to Fast EM $\Pi$GDM method.
    }
    \label{fig:fid_on_runtime}
    \vspace{-5pt}
\end{figure}
\begin{figure*}[t]
    \centering
    \includegraphics[width=\textwidth]{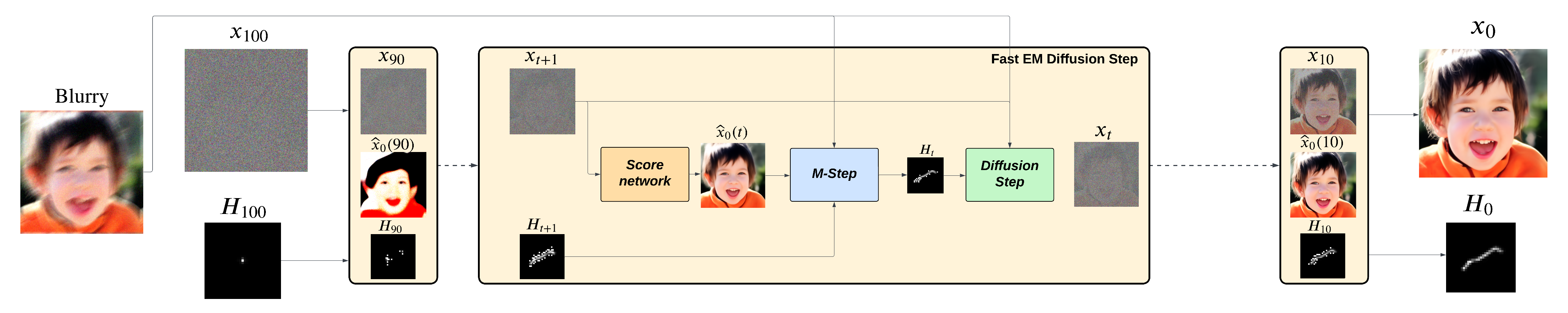}
    \caption{Overview of the method and evolution of the current estimates. We start with random noise and apply the diffusion process. The blurry image intervenes both for the guidance and for the M-step which estimates the blur kernel.}
    \vspace{-5pt}
\end{figure*}
This growing interest in diverse restoration is motivated by the impressive perceptual quality obtained by such methods. In particular, diffusion models that were first introduced for image synthesis tasks~\cite{ho_denoising_2020, song_denoising_2021, ho_classifier-free_2022} are now used for a large diversity of tasks such as inverse problem solving~\cite{kawar_denoising_2022, chung_diffusion_2023, song_pseudoinverse-guided_2023}.  
In the field of blind deconvolution, it is common to use Bayesian methods to jointly estimate the blur kernel and the restored image~\cite{perrone_total_2014, anger_blind_2019, luo_deep_2022, ren_neural_2020}. The kernel estimation highly relies on the restoration method that is used and it generally requires the restoration method to produce a sharp image. To do so, image regularizations such as TV, $\ell_0$ on the gradient can be used but they tend to over-sharpen the restored image leading to unpleasant results.  
Even with the sharp and blurry pairs, it is not easy to estimate any type of blur kernels without efficient regularization. Common regularization on the kernels are the $\ell_1$ norm~\cite{chung_parallel_2023}, positivity, the sum to one constraint, and in some cases Gaussian constraints~\cite{bell-kligler_blind_2019}. Some recent works also use deep neural networks such as normalizing flows to parameterize the kernels~\cite{liang21fkp}.
Motivated by the impressive quality of diffusion models for both estimated conditional distribution and returning high-quality images, it is natural to believe that they could be used in the context of kernel estimation. Also, a pioneer work~\cite{chung_parallel_2023} that combines parallel diffusion models for the kernel and image exhibits impressive results. Estimating the kernel and image is jointly done in the diffusion process using gradient descent on the forward model. 
Similarly, methods based on Monte Carlo sampling proposed parameters estimation derived from the Expectation-Maximization (EM) algorithm~\cite{guo_agem_2019, gao_deepgem_2021}, or the SAPG algorithm~\cite{Vidal2020, Fort2019}.
Those methods are very efficient but Monte Carlo sampling is time-consuming. Also, the problem of kernel estimation is a complex problem so those methods highly depend on the regularization imposed in the \textit{M-step} of the EM algorithm.
\newline
\newline
Motivated by the efficiency of diffusion models, we propose a diffusion model that solves the maximum a-posteriori estimator for blind deconvolution. Derived from the classical Expectation-Maximization algorithm, our model alternately estimates the expected value of the log-likelihood using samples drawn from a diffusion model and maximizes this quantity using half-quadratic splitting. In addition, we also propose a novel kernel regularization in a Plug \& Play fashion. Finally, we proposed a fast version of our algorithm to facilitate the use of our method in real-world scenarios. Our experiments show that our proposed solution improves both in terms of fidelity and computational efficiency pushing the Pareto optimal curve further to the origin (Figure~\ref{fig:fid_on_runtime}).

\section{Background}

Let us suppose that our deblurring problem fits the classical inverse problem formulation:
\begin{equation}
\label{equ:inverse-pbm}
    y = Hx + n \quad \text{with} \quad n \sim \mathcal{N}(0, \sigma^2)
\end{equation}
where $x$ is the clean image we want to estimate, $y$ is the blurry and noisy image and $H$ is the degradation operator, a convolution operator in the case of deconvolution.
We suppose that we are in the real-world case where we only have access to the blurry image $y$ and the noise level $\sigma$ to reconstruct both the clean image and the blur kernel $H$. 
In such a setting, a common approach to estimate the blur kernel is to compute the marginalized maximum a-posteriori (MAP) estimator of the inverse problem described in Equation~\eqref{equ:inverse-pbm}:
\begin{align}
    \label{equ:likelihood}
    & H_{MAP} = \arg\max_H p(H|y) = \arg\max_H p(y|H)p(H) \\
    & = \arg\max_H \left[\log{\left(\int p(y|H,x)p(x)dx\right)} + \log(p(H)) \right], \nonumber
\end{align}
with $p(x)$ a natural image prior, $p(y|H,x)$ the likelihood of the blurry image and $p(H)$ the kernel's prior distribution.
This MAP estimator cannot be solved easily since the marginalization in the clean image $x$ is not tractable. Expectation-Maximization (EM)~\cite{Dempster1977, McLachlan2008} is an iterative algorithm that computes the MAP estimator for the parameters of a statistical model ($H$ in our case). It is very convenient when the model contains unobserved or missing data. The EM algorithm consists of two main steps. An \textit{E-step} that computes the expected log-likelihood given the current model parameter estimates and an \textit{M-step}, that maximizes this expected log-likelihood to update the estimated parameters. The whole algorithm alternates between the \textit{E-step} and \textit{M-step} until convergence. In the case of deblurring, the parameter we want to estimate is the blur kernel $H$ and our unobserved data are the clean images associated with the blurry image $y$ and the estimated blur kernel $H$. The EM algorithm can be summarized as follows in such setting:
\newline
\noindent\textit{E-Step:}
\begin{align}
\label{equ:e_step}
    Q(H,H_l) & = E_{x\sim p(x|y,H_l)}[\log(p(y|x,H)) + \log(p(x))]
\end{align}
\noindent\textit{M-Step:} 
\begin{align}
\label{equ:m_step}
    H_{l+1} &= \arg\max_H \left[Q(H,H_l) + \log(p(H))\right]
\end{align}
This formulation is very convenient but in many applications (including blind deblurring), the expected log-likelihood in Equation~\eqref{equ:e_step} cannot be computed explicitly, and even taking posterior samples $x \sim p(x|y,H_l)$ is challenging.
Our method proposes to approximate the expectation in the \textit{E-step} by an empirical mean in Monte-Carlo EM fashion~\cite{Wei1990AMC} and to use a diffusion model to obtain posterior samples.
\newline
\newline
\textbf{Diffusion models for posterior sampling:} To learn $p(x_0)$ the distribution of the data, diffusion models define a family of distributions $p(x_t)$ by gradually adding Gaussian noise of variance $\beta(t)$ to samples of $p(x_0)$ until the distribution $p(x_T)$ reduces to a standard Gaussian with zero mean. 
For discrete timesteps $t \in \llbracket 0, T \rrbracket$, we can define a Markov transition kernel $p(x_t|x_{t-1}) = \mathcal{N}(x_t; \sqrt{1-\beta(t)}x_{t-1}, \beta(t) I)$ between two consecutive discrete timestamps.
In the general continuous case, \cite{song_score-based_2021} described the forward noising process with the following stochastic differentiable equation (SDE) :
\begin{equation}
    d{x_t} = -\frac{\beta(t)}{2} {x_t} dt + \sqrt{\beta(t)}dw 
\end{equation}
where $w(t)$ is the d-dimensional Wiener process. The reverse SDE of this process~\cite{anderson_reverse-time_1982} can be written as:
\begin{equation}
\label{equ:reverse-SDE}
    d{x_t} = [-\frac{\beta(t)}{2}{x_t} - \beta(t)\nabla_{{x_t}} \log \pi({x_t}) ]dt + \sqrt{\beta(t)}d\bar{w}
\end{equation}
with $dt$ corresponding to time running backwards and $d\bar{w}$ to the standard Wiener process running backwards. 
In the case of inverse problems, we want to use diffusion models to generate the posterior distribution $\pi(x_t) = p({x_t}|y, H)$. Using Bayes' rule Equation~\eqref{equ:reverse-SDE} becomes:
\begin{dmath}
\label{equ:reverse-SDE-inv-pbm}
    d{x_t} =  \left[-\frac{\beta(t)}{2}{x_t} - \beta(t)\left(\nabla_{{x_t}} \log p({x_t}) + \textcolor{blue}{\nabla_{{x_t}} \log p(y|{x_t}, H)}\right) \right]dt + \sqrt{\beta(t)}d\bar{w}
\end{dmath}
The main problem behind this equation is that in inverse problems, we have a relation between $y$ and $x_0$ but not between $x_t$ and $y$. Marginalizing in $x_0$, we obtain:
\begin{equation}\label{eq:marginalize}
    p(y|x_t) = \int p(y|x_0)p(x_0|x_t)dx_0
\end{equation}
that is intractable. 
The main challenge of non-blind diffusion for posterior sampling is to compute or approximate this integral. In our work, we conduct experiments with DPS~\cite{chung_diffusion_2023} and $\Pi$GDM~\cite{song_pseudoinverse-guided_2023} that use different approximations for this integral. Both approximations are based on the mean of $p(x_0|x_t)$, namely:
\[\label{eq:x0hat}
\xZeroHat := E[x_0|x_t].\]
DPS approximates $p(x_0|x_t)$ by a delta function
\begin{equation}\label{eq:dps-approx}
    p(x_0|x_t) \approx \delta_{\xZeroHat}(x_0)
\end{equation}
whereas $\Pi$GDM approximates $p(x_0|x_t)$ by a Gaussian distribution
\begin{equation}\label{eq:pigdm-approx}
    p(x_0|x_t)\approx \mathcal{N}(x_0 | \xZeroHat, r_t^2)
\end{equation}
with $r_t$ a hyper-parameter.
Both approximations allow us to solve the marginal in Equation~\eqref{eq:marginalize} analytically and obtain explicit expressions for $\nabla_{{x_t}} \log p(y|{x_t})$ as detailed below.
\newline
\newline
As a recall, one property of diffusion models is that we can express the noisy measurement $x_t$ in the forward model using the original sample $x_0$:
\begin{equation}
    x_t = \sqrt{\bar{\alpha}_t} x_0 + \sqrt{1-\bar{\alpha}_t}\epsilon
\end{equation}
with $\alpha_t = 1 - \beta_t$ and $\bar{\alpha}_t = \prod\limits_{i=1}^t{\alpha_i}$.
\newline
Using a noise predictor $\epsilon(x_t, t)$, we can thus estimate $\xZeroHat = E[x_0|x_t]$ at each step $t$ using:
\begin{equation}
    \label{equ:x_0_hat}
    \xZeroHat = \frac{1}{\sqrt{\bar{\alpha}_t}}(x_t - \sqrt{1-\bar{\alpha}_t}\epsilon(x_t, t)).
\end{equation}
Equivalently, we can use a score network $s(x_t, t)$ using Tweedie's identity:
\begin{equation}\label{eq:Tweedie}
    s(x_t, t) = \nabla_{x_t} \log p(x_t)= -\frac{1}{\sqrt{1-\bar{\alpha}_t}}\epsilon(x_t, t).
\end{equation}
Using DDPM~\cite{ho_denoising_2020} to discretize the unconditional reverse diffusion process~\eqref{equ:reverse-SDE} we obtain the update rule
\begin{equation}\label{eq:DDPM-update}
    x_{t-1} = \frac{1}{\sqrt{\alpha_t}}\left( x_t + \beta_t s(x_t, t)\right) + \tilde{\sigma}_t \mathcal{N}(0,I)
\end{equation}
where $ \tilde{\sigma}_t = \sqrt{\beta_t} \text{ or } \sqrt{\frac{(1-\bar\alpha_{t-1})}{1-\bar{\alpha}_t}\beta_t}$. To simulate the \emph{conditional} reverse diffusion process~\eqref{equ:reverse-SDE-inv-pbm}, we just have to add the likelihood term to the score
\begin{dmath}\label{eq:conditional-DDPM-update}
    x_{t-1} = \frac{1}{\sqrt{\alpha_t}}\left( x_t + \beta_t \left[s(x_t, t) + \nabla_{x_t} \log p(y|x_t)\right] \right) + \tilde{\sigma}_t \mathcal{N}(0,I) 
\end{dmath}
Using Equation~\eqref{equ:x_0_hat}, the DPS~\cite{chung_diffusion_2023} approximation for $p(x_0|x_t)$ leads to the following formula for the gradient of the log-likelihood:
\begin{equation}
    \label{equ:dps_grad}
    \nabla_{x_t} \log p(y|x_t) = -\frac{1}{\sigma^2}\nabla_{x_t} \|y - H\xZeroHat\|_2^2
\end{equation}
Similarly, the $\Pi$GDM~\cite{song_pseudoinverse-guided_2023} approximation leads to the following gradient for the log-likelihood:
\begin{align}
    \label{equ:pigdm_grad}
    & \nabla_{x_t} \log p(y|x_t) = \\
    & \left((y - H\xZeroHat)^T(r_t^2 H H^T + \sigma^2 I)^{-1}H \left(\frac{\partial \xZeroHat}{\partial x_t}\right) \right)^T \nonumber
\end{align}
DPS and $\Pi$GDM derive different guidance terms for the inverse problem. While the DPS approximation leads to a gradient that is easily implemented for any degradation operator $H$ using automatic differentiation, the $\Pi$GDM approximated gradient of Equation~\eqref{equ:pigdm_grad} is much more complex to estimate for a general operator $H$ because it requires the computation of its pseudo-inverse. 
On the other hand, the $\Pi$GDM approximation is more precise and thus leads to stronger guidance which is very important for kernel estimation. 
We summarize in Algorithm~\ref{alg:diff_inv_pbm} the diffusion process for inverse problems when the degradation operator $H$ is known. This case covers both DPS and $\Pi$GDM. The pseudo-code is written using DDPM but is not limited to this particular diffusion scheme. 
To compensate for the fact that the first estimations of $x_t$ are uncertain, it is common to set $\zeta_t = \sqrt{\bar{\alpha}_t}$, instead of the theoretical $\zeta_t=1$.

\section{Method}
Our method proposes to solve the MAP of the blur kernel from a blurry and potentially noisy image. We estimate the MAP estimator in an EM fashion. Iteratively, we first draw samples from the posterior distribution knowing the current kernel estimate using a diffusion model. It corresponds to the \textit{E-step} of the EM algorithm. Then, we update our estimated kernel with the \textit{M-step} by maximizing the expected log-likelihood on the previously computed samples. To efficiently model the kernels' distribution, we use a Plug \& Play kernel denoiser to regularize our MAP estimator.
\subsection{E-step: Non-blind diffusion}
The \textit{E-step} of the EM algorithm consists in evaluating the expectation from Equation~\eqref{equ:e_step}. Instead of computing its exact value, we propose to approximate it using random samples in a Monte-Carlo EM fashion. To draw the random samples, we use a non-blind diffusion model. Since the diffusion model targets $p(x|y, H_l)$, sampling several images leads to a good approximation of the expectation. The number $n$ of samples used to approximate the expectation is a hyperparameter of the method. Having many samples leads to a slow but accurate estimation while having only one sample is equivalent to the Stochastic EM algorithm~\cite{nielson2000sEM}. In practice, the \textit{E-step} reduces to:\\
    \emph{Drawing samples}
    \begin{equation}\label{eq:Estep1}
    \xb = (x^1,...,x^n) \sim p(x_0|y,H_l)\end{equation}
    and \emph{updating}
    \begin{equation}\label{eq:Estep2}
    \widehat{Q}(H,H_l) = \frac{1}{n}\sum_{i=1}^n{\log(p(y|x^i,H))} .
    \end{equation}
The samples can be drawn by $n$ parallel runs of Algorithm~\ref{alg:diff_inv_pbm}, and the empirical mean $\widehat{Q}(H,H_l) \approx {Q}(H,H_l)$ approaches the expected value in Equation~\eqref{equ:e_step} as $n\to\infty$. Unlike in Equation~\eqref{equ:e_step}, we remove the term in $p(x)$ from $\widehat{Q}(H,H_l)$ here since it does not affect the maximization in the blur kernel $H$.

\subsection{M-step: Kernel estimation}
The \textit{M-step} computes the MAP estimator of the blur kernel using the estimated samples from the \textit{E-step} as measurements.
From equations (1), (4) and (19) this step can be summarized as:
\begin{align}
    \label{equ:M_step_1}
    H_{l+1} & = \arg\max_H \hat{Q}(H,H_l) + \log(p(H)) \\
    H_{l+1} & = \arg\min_H \frac{1}{2 n \sigma^2}\sum_{i=1}^n{\|y - H x^i\|_2^2} + \lambda \Phi(H) \label{equ:M_step_2} 
\end{align}
where~\eqref{equ:M_step_2} is obtained using Equation~\eqref{equ:inverse-pbm} and~\eqref{eq:Estep2}.
Common choices for $\Phi(.)$ are $\ell_2$ or $\ell_1$ regularizations on top of the simplex constraints on the blur kernel (non-negative values that add up to one).
Despite being quite efficient when the blurry image does not have noise, they generally fail to provide good quality results when the noise increases. On the other side, Plug \& Play regularizations have become more and more popular for many image restoration tasks. By training a deep denoiser on Gaussian denoising, one can obtain a powerful regularization in the domain on which the denoiser was trained. Generally, we train the denoiser on a dataset of natural images leading to a regularization on natural images. Here, we propose to train a denoiser on a dataset of blur kernels to build a Plug \& Play regularization for the blur kernels. We observed that this approach leads to a kernel estimation algorithm that is more efficient and robust to noise, see Figure~\ref{fig:reg_compar}. 
To solve Equation~\eqref{equ:M_step_2}, we use the Half-Quadratic Splitting (HQS) optimization scheme:
\begin{align}
\label{equ:HQS-ker-1}
    \notag
    Z_{j+1} = \arg\min_Z & \frac{1}{2\sigma^2 n}\sum_{i=1}^n{\|Z x^i -y\|_2^2} \\
    & + \frac{\beta}{2}\|Z-K_j\|_2^2 \\
\label{equ:HQS-ker-2}
    K_{j+1} = \arg\min_K & \lambda \Phi(K) + \frac{\beta}{2}\|K-Z_{j+1}\|_2^2
\end{align}
For the deconvolution problem, Equation~\eqref{equ:HQS-ker-1} can easily be solved in the Fourier domain (more details on the computations can be found in Appendix B).
Equation~\eqref{equ:HQS-ker-2} corresponds to the regularization step. It corresponds to the MAP estimator of a Gaussian denoising problem on the variable $Z_{j+1}$. The main idea behind Plug \& Play regularization is to replace this regularization step with a pre-trained denoiser $\mathcal{D}$ Mean Squared Error (MSE) loss. This substitution can be done thanks to the close relationship that exists between the MAP and the MMSE estimator of a Gaussian denoising problem~\cite{Gribonval2011}. Eventually, the \textit{M-step} consists of the following iterations:
\begin{align}
    \label{equ:m_step_fin_1}
    & Z_{j+1} = \mathcal{F}^{-1}\left(\frac{\mathcal{F}(y)\sum_{i=1}^n{\overline{\mathcal{F}(x^i)} +  n \beta \sigma^2 \mathcal{F}(K_j)}}{\sum_{i=1}^n{\mathcal{F}(x^i)\overline{\mathcal{F}(x^i)}+n \beta \sigma^2 }}\right)  \\
    \label{equ:m_step_fin_2}
    & K_{j+1} = \mathcal{D}_{\sqrt{\lambda/\beta}}(Z_{j+1}).
\end{align}
While complex decreasing schemes for $\beta$ are often used to help HQS converge~\cite{zhang2021dpir}, we observed that using a constant $\beta$ was sufficient in our case. 
For the denoiser architecture, we use a simple DnCNN~\cite{zhang_beyond_2017} with 5 blocks and 32 channels. In addition to the noisy kernel, we also give the noise level as an extra channel to the network to control the denoising intensity. 
Eventually, the complete Diffusion EM algorithm alternates between sampling from the non-blind diffusion model and the HQS algorithm for the kernel estimation. In all our experiments, we use $L=10$ EM iterations. See Algorithm A.1 in the supplementary.
\setlength{\textfloatsep}{5pt}
\begin{algorithm}[t] \small
    \caption{Diffusion model for deblurring}\label{alg:diff_inv_pbm}
    \begin{algorithmic}
        \Require $y, \sigma, H, T, (\zeta_t)_t$
        \Ensure A posterior sample $x_{0} \sim p(x_{0}|y,H)$
        \State $x_T \gets \mathcal{N}(0,I)$
        \For{$t=T$ \textbf{to} $1$}
            \State $\widehat{\epsilon} \gets \epsilon(x_t, t)$
            \State $\widehat{x}_0 = \frac{1}{\sqrt{\bar{\alpha}_t}}(x_t - \sqrt{1-\bar{\alpha}_t}\widehat{\epsilon})$
            \State \mycomment{DPS or $\Pi$GDM approx. using $\hat{x}_0$}
            \State $g \gets \nabla_{x_t} \log p(y|x_t, H) $ \Comment{Equation~\eqref{equ:dps_grad}~or~\eqref{equ:pigdm_grad}}
            \State \mycomment{Compute conditional score $s=\nabla_{x_t}\log p(x_t|y,H)$}
            \State $s \gets \zeta_{t} g - \frac{1}{\sqrt{1-\bar\alpha_t}}\hat{\epsilon}$ \Comment{Bayes rule and Tweedie}
            \State \mycomment{DDPM update rule}
            \State $ z \gets \mathcal{N}(0,I)$
            \State ${x}_{t-1} \gets \frac{1}{\sqrt{\alpha_t}}\left( x_t + \beta_t s\right) + \tilde\sigma_t z$
        \EndFor
        \State \Return $x_0$
    \end{algorithmic}
\end{algorithm}
\begin{algorithm}[t]\small
    \caption{Fast EM DPS / $\Pi$GDM}
    \label{alg:fast_diffEM}
    \begin{algorithmic}
        \Require $y, \sigma, H_T, T$
        \Ensure $H \approx \arg\min_{H} p(y|H)$ and 
        $x_0^i \sim p(x_0|y,H)$

        \State $\boldsymbol{x_T} \gets (\mathcal{N}(0,I), ..., \mathcal{N}(0,I)) \in (\mathbb{R}^{h*w*3})^n$ 
        \For{$t=T$ \textbf{to} $1$}
            \State $\boldsymbol{\widehat{\epsilon}} \gets \epsilon(\boldsymbol{x}_t, t)$
            \State $\boldsymbol{\widehat{x}}_0 = \frac{1}{\sqrt{\bar{\alpha}_t}}(\boldsymbol{x}_t - \sqrt{1-\bar{\alpha}_t}\boldsymbol{\widehat{\epsilon}})$
            \State $\color{blue}{H_{t-1} = \textit{M-step}(y, \boldsymbol{\widehat{x}_0}, \sigma)}$ \Comment{Iterate~\eqref{equ:m_step_fin_1}~and~\eqref{equ:m_step_fin_2}}
            
            \State \mycomment{DPS or $\Pi$GDM approx. using $\hat{x}_0$}
            \State $\boldsymbol{g} \gets \nabla_{\boldsymbol{x}_t} \log p(y|\boldsymbol{x}_t, H_{t-1}) $ \Comment{Equation~\eqref{equ:dps_grad}~or~\eqref{equ:pigdm_grad}}

            \State \mycomment{Compute conditional score $s=\nabla_{x_t}\log p(x_t|y,H)$}
            \State $\boldsymbol{s} \gets  \zeta_{t} \boldsymbol{g} - \frac{1}{\sqrt{1-\bar\alpha_t}}\boldsymbol{\hat{\epsilon}}$ \Comment{Bayes rule and Tweedie}

            \State \mycomment{DDPM update rule}
            \State $ \boldsymbol{z} \gets (\mathcal{N}(0,I), ..., \mathcal{N}(0,I)) \in (\mathbb{R}^{h*w*3})^n$
            \State $\boldsymbol{x}_{t-1} \gets \frac{1}{\sqrt{\alpha_t}}\left( \boldsymbol{x}_t + \beta_t \boldsymbol{s}\right) + \tilde\sigma_t \boldsymbol{z}$
            
        \EndFor
        \State \Return $\boldsymbol{x_0}$, $H_{0}$
    \end{algorithmic}
\end{algorithm}

\subsection{Fast EM diffusion}
\setlength{\tabcolsep}{6pt}
\begin{table*}[t]
    \centering
    \resizebox{0.9\textwidth}{!}{%
    \begin{tabular}{|c|c|cccc|cc|cc|}
        \hline
         Metric type & & \multicolumn{4}{|c|}{Reference metrics} & \multicolumn{2}{|c|}{No-reference metrics} & \multicolumn{2}{|c|}{Kernel error} \\
         \hline
        $\downarrow$ Method $\backslash$ Metric $\rightarrow$& Time (sec/img) & PSNR $\uparrow$ & SSIM $\uparrow$ & LPIPS $\downarrow$ & FID $\downarrow$ & NIQE $\downarrow$ & BRISQUE $\downarrow$ & MSE kernel  $\downarrow$ &  $\mathcal{L}_{reblur}$ $\downarrow$\\
        \hline\hline
        DPS* & 58sec & 25.81 & 0.76 & 0.34 & 3.46 & 6.28 & 23.52 & \xmark & \xmark  \\
        $\Pi$GDM* & 5sec & 27.65 & 0.81 & 0.34 & 4.50 & 7.49 & 30.32 & \xmark & \xmark \\
        \hline\hline
        Anger $\ell_0$ & 0.73sec & 12.46 & 0.13 & 0.8 & 233.08 & 12.55 & 50.51 & 5.1e-5 & 1.1e-2\\
        Self-Deblur & 1min53sec & 14.53 & 0.15 & 0.69 & 44.83 &14.16 & 49.28 & 3.6e-4 & 3.5e-2\\
        MPRNet & 3.7sec & 19.52 & 0.42 & 0.54 & 21.26 &  7.9 & 25.44 & \xmark & \xmark \\
        Blind DPS & 1min23 & 24.05 & 0.73 & \textbf{0.34} & \textbf{2.66} & \textbf{6.17} & \textbf{20.72} &  3.9e-5 & 5.6e-3\\
        \hline
        EM $\Pi$GDM (n=1) & 1min30sec & 23.4 & 0.71 & 0.43 & 6.05 & 8.81 &   41.19 & 6.1e-5 & 5.3e-3 \\
        EM $\Pi$GDM (n=4) & 2min30sec & 23.21 & 0.71 & 0.4 & 5.43 & 8.23 &  38.02 & 5e-5 & 5.3e-3 \\
        EM $\Pi$GDM (n=16) & 9min10sec & 23.09 & 0.71 & \underline{0.39} & 5.11 & 7.91 &  35.42 & 4.1e-5 & 5.3e-3 \\
        \hline
        Fast EM DPS (n=1)& 1min41 & 24.68 & 0.75 & \textbf{0.34} & \underline{3.23} & \underline{6.34} & \underline{23.03} & \underline{9e-6} & \underline{5.1e-3}\\
        \hline
        Fast EM $\Pi$GDM (n=1) & 9sec & 25.66 & \underline{0.79} & \textbf{0.34} & 4.26 & 7.48 & 30.33 & 1.1e-5 & \underline{5.1e-3} \\
        Fast EM $\Pi$GDM (n=4) & 15sec & \underline{25.74} & \textbf{0.8} & \textbf{0.34} & 4.31 & 7.42 & 30.15 & \textbf{6e-6} & \textbf{5e-3} \\
        Fast EM $\Pi$GDM (n=16) & 55sec & \textbf{25.75} & \textbf{0.8} & \textbf{0.34} & 4.28 & 7.46 & 29.61 & 1.1e-5 & \textbf{5e-3} \\
        
        \hline
    \end{tabular}%
    }
    \caption{Model comparison on FFHQ synthetic dataset. Models with a ``*'' correspond to \emph{non-blind} models used as baselines. Best \emph{blind} models are in \textbf{bold} while second best are \underline{underlined}. Note that baselines do not count for best model rankings.}
    \label{tab:ffhq_test}
\end{table*}
The diffusion EM algorithm requires 
running a diffusion model at each step of the EM algorithm to produce a set of $n$ particles. Executing diffusion models is time-consuming, particularly in cases where inverse problems are addressed using score guidance, as the guidance must be applied to the full-size image, precluding the utilization of acceleration techniques like latent diffusion~\cite{rombach2022high}. Consequently, the diffusion EM algorithm's execution time becomes excessively long, significantly restricting its practical applicability.
\newline
\newline
To bypass this problem, we propose a fast version of diffusion EM that incorporates the \textit{M-step} directly into the diffusion process, thereby reducing the number of required diffusion model runs to just one. 
To do so, we use the $n$ current samples $x_t^i \sim p(x_t|y, H)$ to build an approximation of $Q(H, H_t)$ at each timestep $t$, as follows.
First, we use the current distribution estimates $p(x_0|x_t)$ (Equations~\eqref{eq:dps-approx}~and~\eqref{eq:pigdm-approx} for DPS, resp. $\Pi$GDM approximations) for each timestep $t$ to approximate the posterior $p(x_0|y, H)$ by (discretized) marginalization on $x_t$:
\begin{align}
    p(x_0|H, y) & = \int p(x_0|x_t)p(x_t|y, H) dx_t \\
    & \approx \sum_{i=1}^n{p(x_0|x_t^i)p(x_t^i|y, H)} \\
    & = \frac{1}{n}\sum_{i=1}^n{p(x_0|x_t^i)} =: q_t(x_0|y,H). 
\end{align}
Then, using this approximation, the \textit{E-step} at timestep $t$ of the diffusion process is reformulated as follows:
\begin{align}
    Q(H, H_t) & = E_{x \sim p(x|y, H_t)}[\log(p(y|x,H)] \\
    & \approx E_{x \sim q_t(x_0|y,H_t)}[\log(p(y|x,H)]
\end{align}
Since the distribution $q_t(x_0|y,H)$ progressively converges to the distribution $p(x_0|y, H)$ as $t\to 0$, we have a finer and finer estimation of the expected log-likelihood and thus, the blur kernel, through the iterations. 
\newline
Finally, the \textit{E-step} reduces in the case of the DPS approximation~\eqref{eq:dps-approx} to:
\begin{align}
    \widehat{Q}(H, H_t) & =  E_{x \sim q_t(x_0|y,H_t)}[\log(p(y|x,H)] \\
    \label{eq:Mstep-dps}
    & = \frac{-1}{2\sigma^2 n} \sum_{i=1}^n{\|H\widehat{x}_0^i(t) - y\|_2^2}.
\end{align}
In this case, the \textit{M-step} is equivalent to the classical diffusion EM \textit{M-step} of Equation~\eqref{equ:M_step_2} but applied in the current estimate $\widehat{x}_0^i(t)$ instead of the real sample $x^i$.
In the case of the $\Pi$GDM approximation~\eqref{eq:pigdm-approx}, we have:
\begin{align}
\label{equ:m-step-pigdm}
    \widehat{Q}(H, H_t) = \frac{-1}{2\sigma^2 n}\sum_{i=1}^n{ E_{x \sim \mathcal{N}( \widehat{x}_0^i(t), r_t^2)}[\|Hx-y\|_2^2]}.
\end{align}
The computations for the \textit{M-step} in that case are left in Appendix D.
Eventually, the only difference between the fast EM diffusion algorithm and a classical non-blind diffusion model is that we first estimate the blur kernel before applying the guidance. Our algorithm demonstrates comparable computational efficiency to non-blind diffusion algorithms, as the computation of the \textit{M-step} negligibly impacts the overall diffusion process. The algorithm's pseudo-code can be found in Algorithm~\ref{alg:fast_diffEM}. Note that in the pseudo-code, the $n$ particles are treated as a batch directly in the $x_t$. 
To point out this difference, all the variables that are seen as a batch are written in bold.
\setlength{\tabcolsep}{0.1 pt}
\begin{figure*}[t]
\centering
\resizebox{1.95\columnwidth}{!}{%
\begin{tabular}{c|cccccc|c}
     LR & Anger $\ell_0$ & Self-Deblur & MPRNet & BlindDPS & Ours + DPS & Ours + $\Pi$GDM & HR\\
    \includegraphics[width=0.125\linewidth]{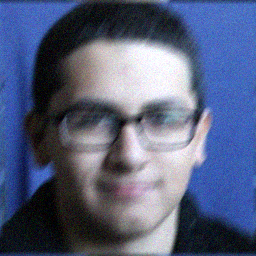} &
    \includegraphics[width=0.125\linewidth]{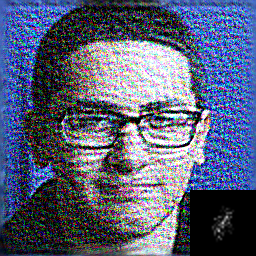} &
    \includegraphics[width=0.125\linewidth]{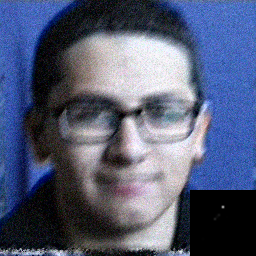} &
    \includegraphics[width=0.125\linewidth]{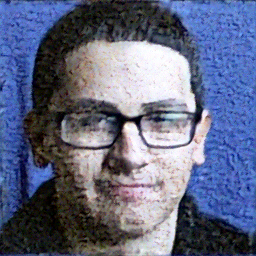} &
    \includegraphics[width=0.125\linewidth]{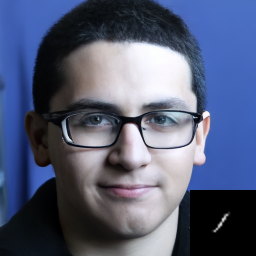} &
    \includegraphics[width=0.125\linewidth]{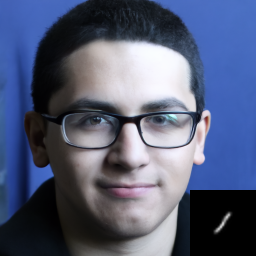} &
    \includegraphics[width=0.125\linewidth]{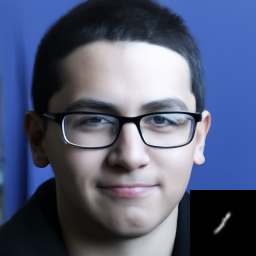} &
    \includegraphics[width=0.125\linewidth]{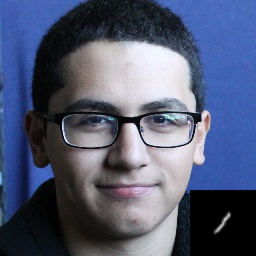} \\
    \includegraphics[width=0.125\linewidth]{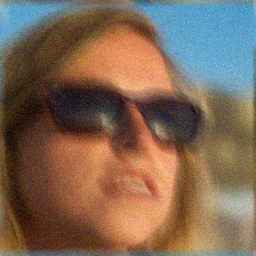} &
    \includegraphics[width=0.125\linewidth]{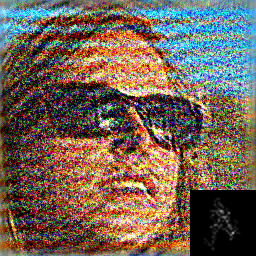} &
    \includegraphics[width=0.125\linewidth]{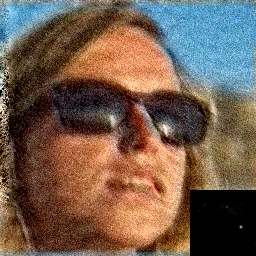} &
    \includegraphics[width=0.125\linewidth]{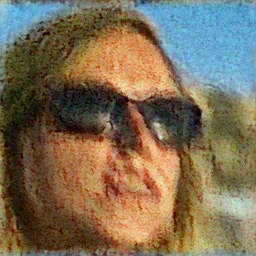} &
    \includegraphics[width=0.125\linewidth]{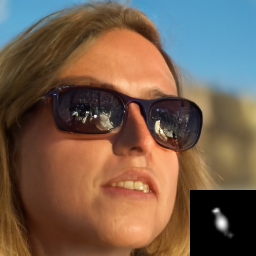} &
    \includegraphics[width=0.125\linewidth]{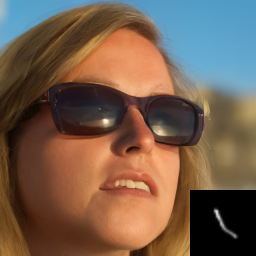} &
    \includegraphics[width=0.125\linewidth]{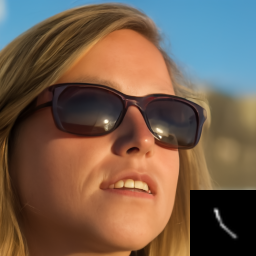} &
    \includegraphics[width=0.125\linewidth]{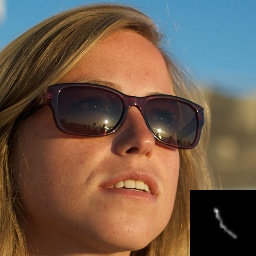} \\
    \includegraphics[width=0.125\linewidth]{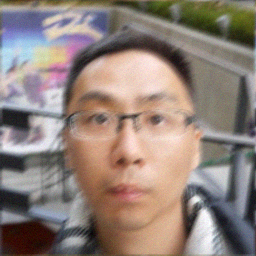} &
    \includegraphics[width=0.125\linewidth]{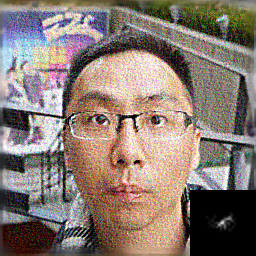} &
    \includegraphics[width=0.125\linewidth]{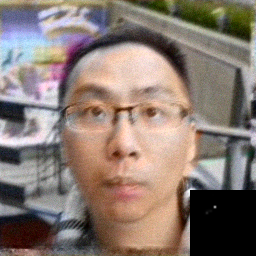} &
    \includegraphics[width=0.125\linewidth]{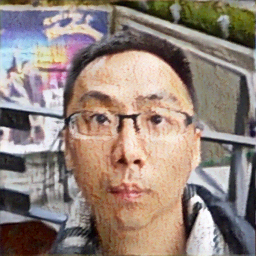} &
    \includegraphics[width=0.125\linewidth]{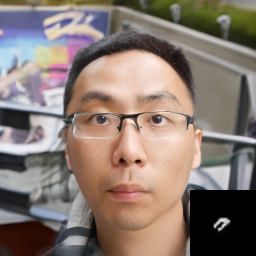} &
    \includegraphics[width=0.125\linewidth]{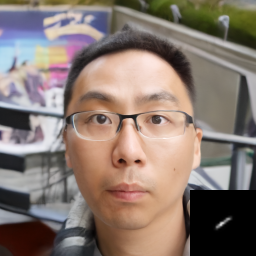} &
    \includegraphics[width=0.125\linewidth]{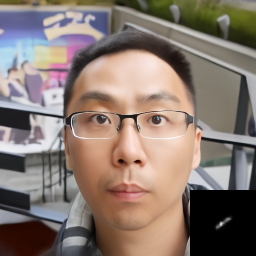} &
    \includegraphics[width=0.125\linewidth]{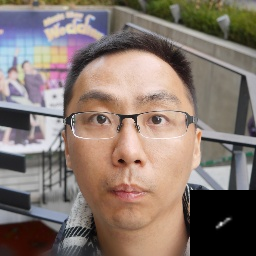}
\end{tabular}%
}
\caption{Visual comparison of the different models on a degraded version of the FFHQ 256x256 dataset. Ours correspond to Fast EM.}
\vspace{-10pt}
\label{fig:visual_res_1}
\end{figure*}

\section{Experiments}

\subsection{Experimental settings}
%
We test our algorithm on the first 1000 validation images of the widely used FFHQ~\cite{ffhqdataset} 256x256 dataset that we degrade with random motion blur kernels computed using~\cite{camera_shake} and random Gaussian noise with noise level $\sigma\in \{5, 10, 20\}$. We also provide some results on DIV2K~\cite{Agustsson_2017_CVPR_Workshops} dataset.
To achieve a fair comparison, we use the code and pre-trained weights provided by the authors of Blind DPS. For $\Pi$GDM, there is no public code so we re-implemented the model using the Blind DPS code backbone. In our experiments, we observed that DPS needs more iterations to properly converge in comparison to $\Pi$GDM. Indeed, the DPS run needs 1000 iterations while we only use 100 iterations for $\Pi$GDM.
%
For the kernel estimation, we use a bias-free FFDNet~\cite{ffdnet} denoiser trained on a dataset of motion blur kernels for the Plug \& Play regularization. At test time, the \textit{M-step} consists of 10 HQS iterations with hyper-parameters $\lambda=1$ and $\beta=1e5$. 
We provide experiments with different numbers of particles for both the Diffusion EM algorithm and the Fast diffusion EM algorithm. We use $n \in \{1, 4, 16\}$. 
All the models are evaluated on a single A100 GPU.

\subsection{Compared methods}

To test the efficiency of our method, we compare it to state-of-the-art models for deconvolution. We chose to compare against both optimization-based methods, deep learning approaches, and diffusion models to cover all the existing approaches. 
More specifically, we compare our method to~\cite{anger_blind_2019} which is a MAP-based method for kernel estimation that uses $\ell_0$ norm on the gradient of the image as an image prior and $\ell_2$ norm to regularize the kernel. 
We also compare to self-deblur~\cite{ren_neural_2020} which is a blind deconvolution method that provides both image reconstruction and kernel estimation based on Deep Image Prior. 
We provide comparisons with MPRNet~\cite{Zamir_2021_CVPR} which is a multi-scale deep learning architecture design for image restoration problems that has proven its efficiency in deblurring. 
Finally, we compare our kernel estimation methods to Blind DPS~\cite{chung_parallel_2023} which consists of two parallel diffusion models that jointly model the restored image and its corresponding blur kernel. 
We also computed the results of the non-blind model DPS and $\Pi$GDM to highlight the loss of quality between the blind and non-blind models.
For all the methods, we used the source code and pre-trained weights provided by the author.
\subsection{Quantitative results}
\begin{figure}
     \centering
     \begin{subfigure}{0.3\columnwidth}
             \centering
             \includegraphics[width=\columnwidth]{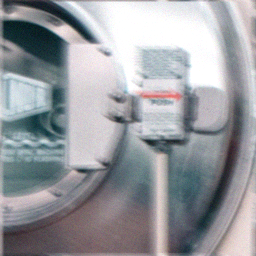}
             \caption{LR}
             \label{fig2:coolcat}
     \end{subfigure}
     \begin{subfigure}{0.3\columnwidth}
             \centering
             \includegraphics[width=\columnwidth]{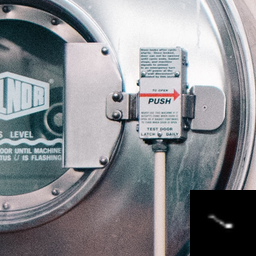}
             \caption{HR}
             \label{fig2:tired}
      \end{subfigure}
      
      \begin{subfigure}{0.3\columnwidth}
             \centering
             \includegraphics[width=\columnwidth]{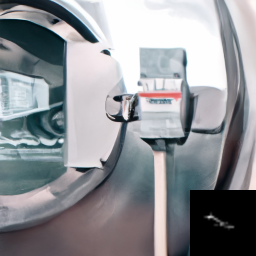}
             \caption{Blind DPS}
             \label{fig2:bossycat}
      \end{subfigure}
      \begin{subfigure}{0.3\columnwidth}
             \centering
             \includegraphics[width=\columnwidth]{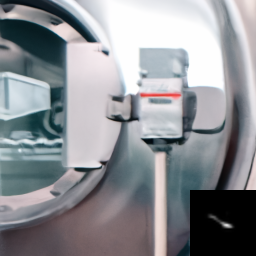}
             \caption{Fast EM DPS}
             \label{fig2:frowningcat}
      \end{subfigure}
      \begin{subfigure}{0.3\columnwidth}
         \centering
         \includegraphics[width=\columnwidth]{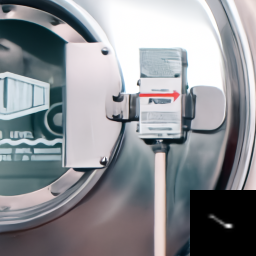}
         \caption{Fast EM $\Pi$GDM}
         \label{fig2:scared}
     \end{subfigure}
\caption{Visual comparison on out-of-distribution images. The score network is trained on FFHQ dataset while we test on DIV2K.}
\vspace{-10pt}
\label{fig:visual_res_2}
\end{figure}

Table~\ref{tab:ffhq_test} shows the results of the different models on FFHQ synthetic dataset.
We compute both classical metrics with full or reduced reference such as PSNR, SSIM~\cite{ssim}, LPIPS~\cite{lpips} and FID~\cite{NIPS2017_8a1d6947}, no-reference metrics to measure perceptual quality such as NIQE~\cite{niqe} and BRISQUE~\cite{brisque} and kernel metrics such as the Mean-Squared Error (MSE) on the reconstructed kernel. We also measure the consistency of the estimated image $\widehat{x}$ and kernel $\widehat{H}$ with the forward model by means of:
\begin{equation}
    \mathcal{L}_{reblur}(y, \widehat{x}, \widehat{H})= \|\widehat{H}\widehat{x} - y\|_2^2 - \sigma^2 M
\end{equation}
where $M=3hw$ is the number of elements in vector $x$.
We observe that classical optimization-based approaches such as Anger $\ell_0$~\cite{anger_blind_2019} and Self-Deblur~\cite{ren_neural_2020} fail to estimate the blur and reconstruct the image efficiently. The main problem with those approaches is that they fail to produce pleasant results in the presence of noise. While Anger $\ell_0$~\cite{anger_blind_2019} produces results with over-sharpened noise, Self-Deblur~\cite{ren_neural_2020} completely fails to both estimate the kernel and deblur the image.
MPRNet produces better results but with artifacts due to the noise, it also fails to recover high-frequency details which is a common problem when using deep-learning models trained on mean-squared error.
\begin{figure}[b]
    \centering
    \includegraphics[trim=15 10 15 40, clip, width=0.65\linewidth]{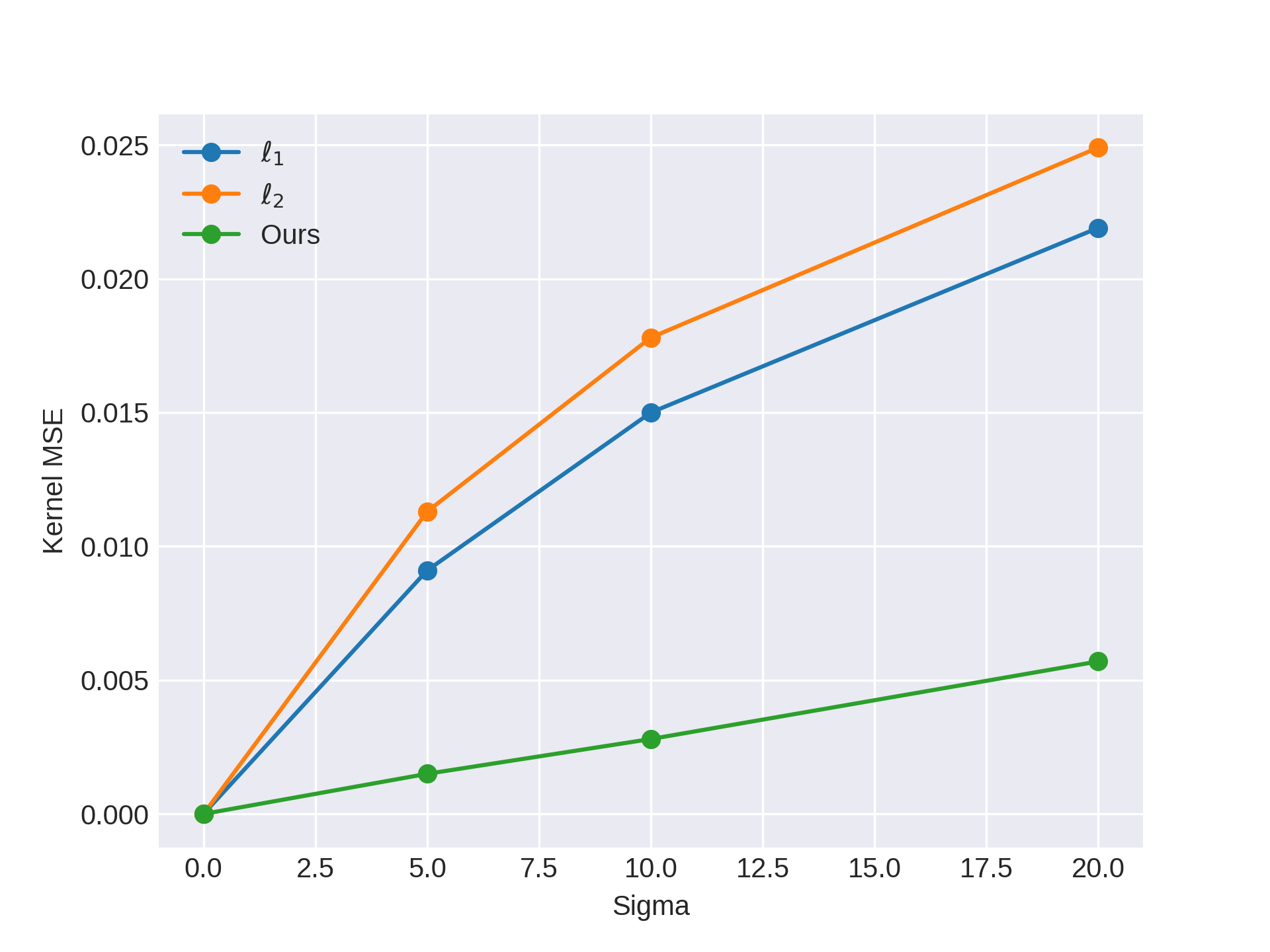}
    \caption{Comparison of the efficiency of the different kernel regularizations depending on the noise level $\sigma\in[0,20]$.
    The vertical axis shows the mean MSE over the whole FFHQ dataset for kernel estimation from a noisy and blurred observation of a known image.}
    \label{fig:reg_compar}
    \vspace{-5pt}
\end{figure}
Diffusion-based models seem to be the most efficient. Blind DPS ranks best among the no-reference perceptual metrics and FID while ranking below our model both for reference metrics and kernel estimation. Figure~\ref{fig:visual_res_1}, shows some example images where we can notice the sharpness and high quality of Blind DPS results. In our experiments, we observed that Blind DPS sometimes fails to efficiently estimate the blur kernel, especially in the presence of noise. We also noticed that on some images Blind DPS was producing sharper results than our model, even with a worst kernel prediction which is surprising since we use the same diffusion model. Yet, the fact that our model has better full-reference metrics and better measurement consistency points out the fact that Blind DPS hallucinates more details. We also conducted experiments on deblurring images from DIV2K dataset while keeping the same FFHQ-trained score model for testing. In that particular case, the prior of the score model does not match the distribution of the test images so the model won’t be able to hallucinate accurate details. Some visual results of those experiments can be found in Figure~\ref{fig:visual_res_2}. Those experiments showed that our model and especially the one based on $\Pi$GDM diffusion produces sharper results. It highlights the fact that
% ...
Blind DPS and DPS, in general, have weaker guidance than $\Pi$GDM, so it requires a more accurate score model which can be a limitation in practice since training a score model on the space of natural images is not an easy task.
During our experiments, we realized that Fast Diffusion EM was both faster and better in terms of quality than Diffusion EM. Indeed, Diffusion EM is sometimes stuck in the no blur solution while we never observed this problem for Fast Diffusion EM. 
In terms of metrics, both Fast EM DPS and Fast EM $\Pi$GDM have better reference metrics than all the other methods, and for any number of particles. We observed better performance and faster runtime with the $\Pi$GDM model, probably because it has stronger guidance, thus, it is easier for the \textit{M-step} to estimate the blur kernel. Fast EM $\Pi$GDM performance in no-reference metrics NIQE and BRISQUE is worse than the other diffusion-based methods: BlindDPS and Fast EM DPS have indeed slightly sharper results, but they are less accurate and less consistent (see the hallucinations of BlindDPS in the second line in Figure~\ref{fig:visual_res_1}).
In terms of runtime, our $\Pi$GDM-based model ranks best among diffusion models but it is significantly slower than MPRNet and Anger $\ell_0$.

\subsection{Ablation studies}
\setlength{\tabcolsep}{6 pt}
\setlength{\textfloatsep}{15pt}
\begin{table}[]
    \centering
    \resizebox{.8\linewidth}{!}{%
    \begin{tabular}{|c|c|c|c|c|}
        \hline
        & n-samples & Runtime & PSNR & PSNR SA \\
        \hline
        \multirow{3}{*}{Diffusion EM} & n=1 & 1min30sec & 23.4 & 23.4 \\
        & n=4 & 2min30sec & 23.21 & 23.43 \\
        & n=16 & 9min10sec & 23.09 & 23.37 \\
        \hline
        \multirow{3}{*}{Fast Diffusion EM} & n=1 & 9sec & 25.66 & 25.66\\
        & n=4 & 15sec & 25.74 & 26.14 \\
        & n=16 & 55sec & 25.75 & 26.16\\
        \hline
    \end{tabular}%
    }
    \caption{Influence of the number of samples used to estimate the \textit{E-step} in Fast EM $\Pi$GDM. The image PSNR is computed on the first image of the batch.}
    \vspace{-10pt}
    \label{tab:n_samples_influence}
\end{table}
In this section, we discuss the efficiency of the different blocks of our algorithm. We first provide some additional results that show the efficiency of the proposed Plug \& Play-based kernel regularization. Next, we study the influence of the number of samples used to estimate the \textit{E-step} on the quality of the final results. 
To compare the efficiency of our regularization, we compared it against the $\ell_1$ and $\ell_2$ regularizations. To do so, we use our FFHQ synthetic dataset and estimate the blur kernel in the non-blind setting where the sharp and blurry images are both known. We compute the MSE of the reconstructed kernel for several noise levels. For all the regularizations, we used the same optimization scheme, HQS, and fine-tuned the hyper-parameters of the regularizations separately. Figure~\ref{fig:reg_compar} shows the obtained results.
We observed that our regularization is significantly better in the presence of noise and the loss of quality between $\sigma=5$ and $\sigma=20$ is very small. 
Finally, we also investigated the influence of the number of samples in our algorithms. We observed in Table~\ref{tab:ffhq_test} and Table~\ref{tab:n_samples_influence} that increasing the number of samples increases the image reconstruction and kernel estimation accuracy. Using all the samples, we can also compute the PSNR on the average of the samples produced by the model. We refer to this metric as the ``PSNR SA'' in Table~\ref{tab:n_samples_influence}. Usually, the PSNR SA gives a higher PSNR than the PSNR on a single image, even if the average image is less sharp. We also observed that in the case of Diffusion EM, increasing the number of samples lowers the PSNR but improves all the other metrics. 
Averaging several samples is also possible with methods such as Blind-DPS, the main difference is that in our approach, all the samples have the same guidance at each diffusion step since we estimate a single kernel for all the samples. In Blind-DPS, all the samples have their respective kernels.

\section{Conclusion}

In this article, we present a novel approach for blind deconvolution based on diffusion models. 
In particular, we designed Diffusion EM, an algorithm based on the Expectation-Maximization algorithm. This algorithm consists of an \textit{E-step}, which approximates the expected value of the log-likelihood using a diffusion model, and an \textit{M-step}, which maximizes this expected log-likelihood with respect to the unknown parameters (the blur kernel).
For the \textit{M-step}, we introduced a novel kernel regularization based on a Plug \& Play denoiser. 
The diffusion EM algorithm is slow since it requires running a diffusion model several times. We propose an acceleration of the algorithm that directly injects the EM iterations into the diffusion process (leveraging the intermediate diffusion steps as approximate posterior samples). We observed that this Fast EM diffusion model reaches better performance than the original diffusion EM algorithm while being significantly faster.
Finally, we demonstrate the efficiency of our approach both quantitatively and visually. We compare our approach to state-of-the-art methods for blind deconvolution and provide several ablation studies that highlight the performance of our regularization and model and give insights into the behavior of the model.
In its current form, our algorithm is limited to deconvolution. Future research will address more general blind deblurring problems~\cite{Debarnot2022,Carbajal2023}.
Faster diffusion models such as latent diffusion~\cite{Rombach_2022_CVPR,Chung2023a} or diffusion bridges~\cite{liu2023i2sb} could also benefit our method.

{\small
\bibliographystyle{ieee_fullname}
\bibliography{egbib}
}

\ifthenelse{\boolean{WithProof}}{\numberwithin{equation}{section}
\numberwithin{algorithm}{section}
\numberwithin{figure}{section}

\newpage\onecolumn
\appendix
\section{Iterative Diffusion EM algorithm}

Algorithm~\ref{alg:diffEM} summarizes the Diffusion EM algorithm described in sections 3.1 and 3.2.
\begin{algorithm}
    \caption{Diffusion EM algorithm}
    \label{alg:diffEM}
    \begin{algorithmic}
        \Require $y, \sigma, H_0, L$,
        \Ensure $H \approx \arg\min_{H} p(y|H)$ and 
        $x_0^i \sim p(x_0|y,H)$

        \For{$l=1$ \textbf{to} $L$}
            \State $\boldsymbol{x} = \textit{E-step}(y, H_{l-1}, \sigma)$ 
            \Comment{$n$ samples from Alg.~1}
            \State $\color{blue}{H_{l} = \textit{M-step}(y, \boldsymbol{x}, \sigma)}$ \Comment{Iterate~(24)~and~(25)}
        \EndFor
        \State \Return $\boldsymbol{x}$, $H_L$
    \end{algorithmic}
\end{algorithm}

\section{M-step computations}\label{sec:Mstep}

\newcommand{\fourier}[1]{{\mathcal{F}(#1)}}

In this section, we derive the computation of the M-step. In particular, we solve Equation~(22) from the main paper:
\begin{align}
Z^* = \arg\min_{Z\in \mathcal{C}} \frac{1}{2\sigma^2 n}\sum_{i=1}^n{\|Z x^i -y\|_2^2} + \frac{\beta}{2}\|Z-H\|_2^2.
\end{align}
with $\mathcal{C}$ the space of convolution operators. 
\newline
In order to account for the fact that $H \in \mathcal{C}$ and $Z_t \in \mathcal{C}$ are convolution operators, we rewrite the same equation in the Fourier domain, where the operators $H$ and $Z$ become diagonal:
\begin{equation}
\fourier{H} = \operatorname{diag}(h(1),\dots,h(d)),
\end{equation}
\begin{equation}
\fourier{Z} = \operatorname{diag}(z(1),\dots,z(d)).
\end{equation}
Re-writing the minimization in the Fourier domain leads to:
\begin{align}
\label{eq:Mstep_appendix_1}
\fourier{Z^*} & = \arg\min_{Z\in\mathcal{C}} \frac{1}{2\sigma^2 n}\sum_{i=1}^n{\|\fourier{Z} \fourier{x^i} -\fourier{y}\|_2^2} + \frac{\beta}{2}\|\fourier{Z}-\fourier{H}\|_2^2 \\
& = \arg\min_{z} \frac{1}{2\sigma^2 n}\sum_{i=1}^n{\sum_{j=1}^d{|z(j) \fourier{x^i}(j) - \fourier{y}(j)|^2}} + \frac{\beta}{2}\sum_{j=1}^d{|z(j) - h(j)|^2}.
\end{align}
It is straightforward that the solution to the problem is also diagonal, thus we have:
\begin{equation}
\fourier{Z^*} = \operatorname{diag}(z^*(1),\dots,z^*(d)).
\end{equation}
Using the first-order condition and the diagonal structure of the problem, we get the following:
\begin{align}
& \frac{1}{\sigma^2 n}\sum_{i=1}^n{\left[z^*(j) \fourier{x^i}(j) -\fourier{y}(j)\right]\overline{\fourier{x^i}(j)}} + \beta (z^*(j) - k(j)) = 0 \\
\Leftrightarrow & z^*(j) \left(\frac{1}{n}\sum_{i=1}^n{|\fourier{x^i}(j)|^2} + \sigma^2 \beta\right) = \fourier{y}(j)\frac{1}{n}\sum_{i=1}^n{\overline{\fourier{x^i}(j)}} + \sigma^2 \beta k(j) \\
\Leftrightarrow & z^*(j) = \frac{\fourier{y}(j)\frac{1}{n}\sum_{i=1}^n{\overline{\fourier{x^i}(j)}} + \sigma^2 \beta k(j)}{\frac{1}{n}\sum_{i=1}^n{|\fourier{x^i}(j)|^2} + \sigma^2 \beta}.
\end{align}

\section{M-step computations with DPS approximation}\label{sec:Mstep-dps}

In this section, we develop the computation of the M-step in Fast EM for DPS. We start from Equation~(32) of the main paper:
\begin{align}\label{eq:proxQ-dps}
    \widehat{Q}(Z, Z_t) = \frac{-1}{2\sigma^2 n}\sum_{i=1}^n{\|Z\widehat{x}_0^i(t)-y\|_2^2]}.
\end{align}
Our goal is to compute:
\begin{align}
    Z^* = arg\min_{Z\in \mathcal{C}} -\widehat{Q}(Z, Z_t) + (\beta/2)\|Z-H\|_2^2.
\end{align}
We can notice that it is similar to Equation~\eqref{eq:Mstep_appendix_1} with $\widehat{x}_0^i(t)$ instead of $x^i$. Thus we have that:
\begin{equation}
    z^*(j) = \frac{\fourier{y}(j)\frac{1}{n}\sum_{i=1}^n{\overline{\mathcal{F}(\widehat{x}_0^i(t))(j)}} + \sigma^2 \beta h(j)}{\frac{1}{n}\sum_{i=1}^n{|\mathcal{F}(\widehat{x}_0^i(t))(j)|^2} + \sigma^2 \beta}.
\end{equation}

\section{M-step computations with $\Pi$GDM approximations}\label{sec:Mstep-pigdm}
In this section, we develop the computation of the M-step in Fast EM for $\Pi$GDM. We start from Equation~(33) of the main paper:
\begin{align}\label{eq:proxQ}
    \widehat{Q}(H, H_t) = \frac{-1}{2\sigma^2 n}\sum_{i=1}^n{ E_{x \sim \mathcal{N}( \widehat{x}_0^i(t), r_t^2)}[\|Hx-y\|_2^2]}.
\end{align}
Our goal is to compute:
\begin{align}
    Z^* = \arg\min_{Z \in \mathcal{C}} -\widehat{Q}(Z, Z_t) + (\beta/2)\|Z-H\|_2^2.
\end{align}
Similarly to Section~\ref{sec:Mstep}, we work with diagonal operators so we have:
\begin{equation}
\fourier{H} = \operatorname{diag}(h(1),\dots,h(d))
\end{equation}
\begin{equation}
\fourier{Z} = \operatorname{diag}(z(1),\dots,z(d)).
\end{equation}
and thus: 
\begin{equation}
\fourier{Z^*} = \operatorname{diag}(z^*(1),\dots,z^*(d)).
\end{equation}
We start by rewriting Equation~\ref{eq:proxQ} in the Fourier domain using the fact that the Fourier transform preserves norms:
\begin{align}
    \fourier{Z^*} = \arg\min_z  \frac{1}{2\sigma^2 n}\sum_{i=1}^n{\sum_{j=1}^d{ E_{x \sim \mathcal{N}(\widehat{x}_0^i(t), r_t^2)}[|z(j)\fourier{x}(j)-\fourier{y}(j)|^2]}} + (\beta/2)\sum_{j=1}^d{|z(j)-h(j)|^2}.
\end{align}
We solve this problem using the first-order condition element by element since the problem is diagonal, the derivation inside the expectancy can be done using Fisher identity \cite[Proposition D.4]{Douc2014}:
\begin{align}
     & \frac{1}{\sigma^2 n}\sum_{i=1}^n{ E_{x \sim \mathcal{N}( \widehat{x}_0^i(t), r_t^2)}[|z(j)\fourier{x}(j)-\fourier{y}(j)|\overline{\fourier{x}(j)}]} + \beta(z(j)-h(j)) = 0 \\
     \Leftrightarrow & z(j)\left[ \frac{1}{n}\sum_{i=1}^n{E_{x \sim \mathcal{N}( \widehat{x}_0^i(t), r_t^2)}[|\fourier{x}(j)|^2]} + \sigma^2\beta \right] = \fourier{y}(j) \frac{1}{n}\sum_{i=1}^n{E_{x \sim \mathcal{N}( \widehat{x}_0^i(t), r_t^2)}[\overline{\fourier{x}(j)}]} + \sigma^2\beta h(j)
\end{align}
Using the fact that the Fourier transform of a white Gaussian noise of variance $\sigma^2$ is a white Gaussian noise of variance $\sigma^2$, the expected values yield:
$$ E_{x \sim \mathcal{N}(\widehat{x}_0, r_t^2)}[|\fourier{x}(j)|^2] = r_t^2 + |\mathcal{F}(\widehat{x}_0)(j)|^2 $$
$$ E_{x \sim \mathcal{N}(\widehat{x}_0, r_t^2)}[\overline{\fourier{x}(j)}] =  \overline{\mathcal{F}(\widehat{x}_0)(j)} $$
So we can conclude that:
\begin{equation}\label{eq:proxQbis}
z^*(j) = \frac{\fourier{y}(j)\frac{1}{n}\sum_{i=1}^n{\overline{\mathcal{F}(\widehat{x}_0^i(t))(j)}} + \sigma^2 \beta h(j)}{\frac{1}{n}\sum_{i=1}^n{|\mathcal{F}(\widehat{x}_0^i(t))(j)|^2}  + r_t^2 + \sigma^2 \beta }
\end{equation}
The main difference with DPS approximation is that we have an extra term in the denominator $r_t^2$.

\section{Additional results}\label{sec:additional_results}

See Figure~\ref{fig:add_res_1}.

\setlength{\tabcolsep}{0.1 pt}
\begin{figure*}[t]
\centering
\resizebox{0.95\columnwidth}{!}{%
\begin{tabular}{c|cccccc|c}
     LR & Anger $\ell_0$ & Self-Deblur & MPRNet & BlindDPS & Ours + DPS & Ours + $\Pi$GDM & HR\\
    \includegraphics[width=0.125\linewidth]{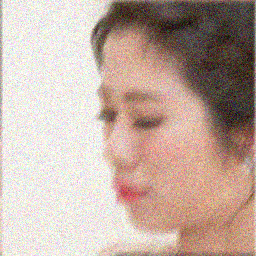} &
    \includegraphics[width=0.125\linewidth]{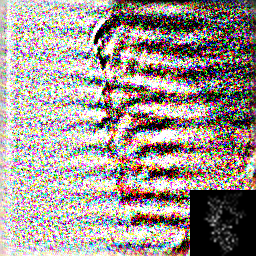} &
    \includegraphics[width=0.125\linewidth]{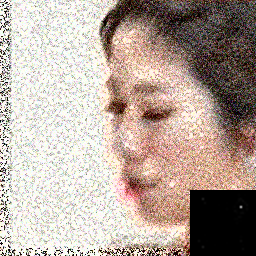} &
    \includegraphics[width=0.125\linewidth]{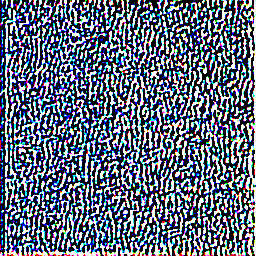} &
    \includegraphics[width=0.125\linewidth]{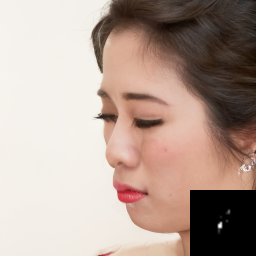} &
    \includegraphics[width=0.125\linewidth]{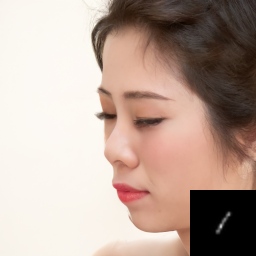} &
    \includegraphics[width=0.125\linewidth]{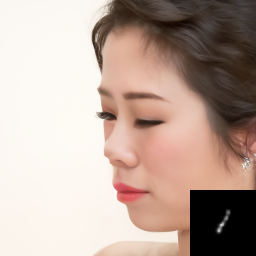} &
    \includegraphics[width=0.125\linewidth]{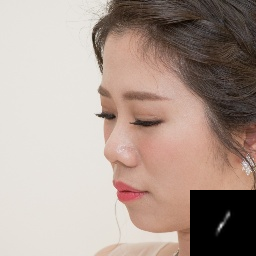} \\
    \includegraphics[width=0.125\linewidth]{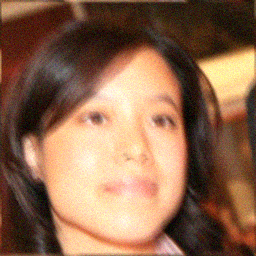} &
    \includegraphics[width=0.125\linewidth]{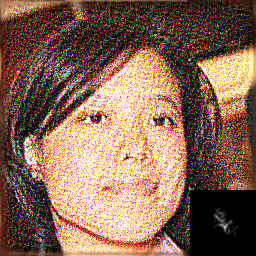} &
    \includegraphics[width=0.125\linewidth]{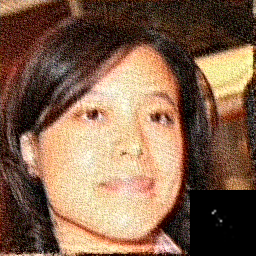} &
    \includegraphics[width=0.125\linewidth]{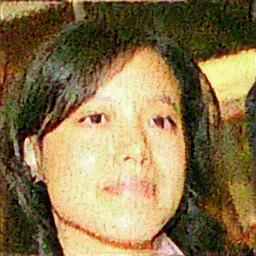} &
    \includegraphics[width=0.125\linewidth]{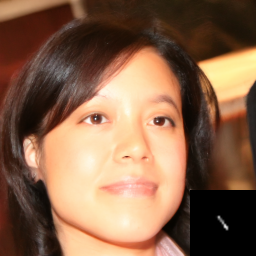} &
    \includegraphics[width=0.125\linewidth]{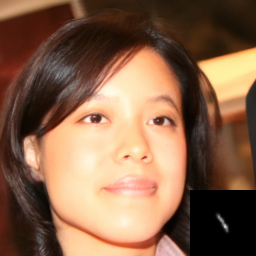} &
    \includegraphics[width=0.125\linewidth]{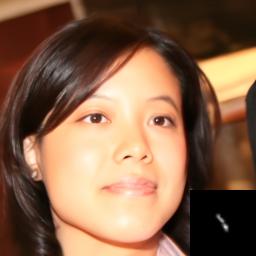} &
    \includegraphics[width=0.125\linewidth]{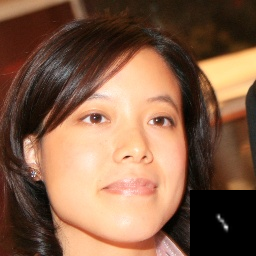} \\
    \includegraphics[width=0.125\linewidth]{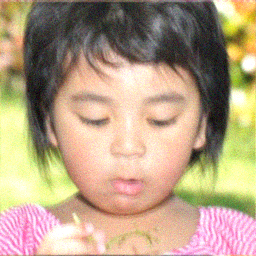} &
    \includegraphics[width=0.125\linewidth]{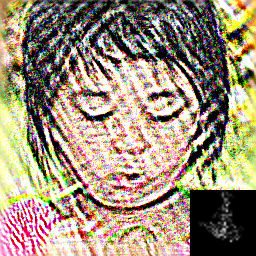} &
    \includegraphics[width=0.125\linewidth]{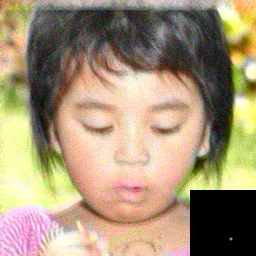} &
    \includegraphics[width=0.125\linewidth]{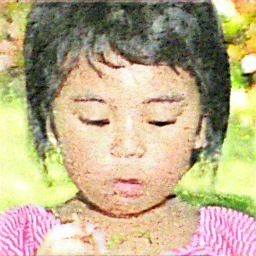} &
    \includegraphics[width=0.125\linewidth]{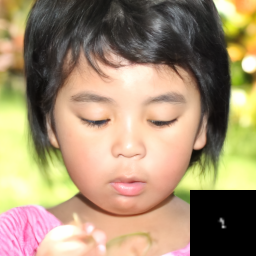} &
    \includegraphics[width=0.125\linewidth]{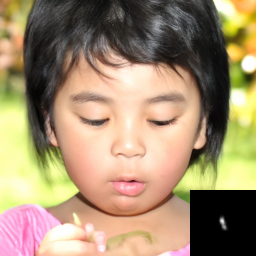} &
    \includegraphics[width=0.125\linewidth]{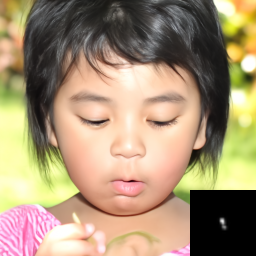} &
    \includegraphics[width=0.125\linewidth]{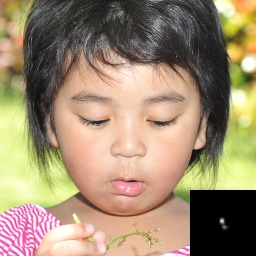} \\
        \includegraphics[width=0.125\linewidth]{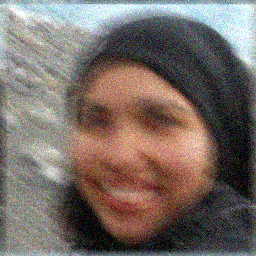} &
    \includegraphics[width=0.125\linewidth]{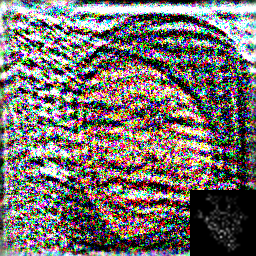} &
    \includegraphics[width=0.125\linewidth]{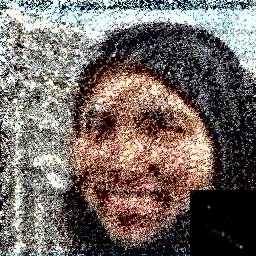} &
    \includegraphics[width=0.125\linewidth]{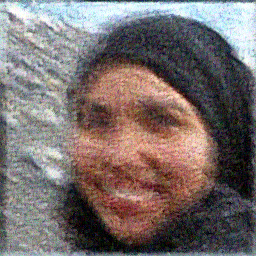} &
    \includegraphics[width=0.125\linewidth]{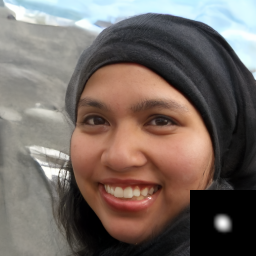} &
    \includegraphics[width=0.125\linewidth]{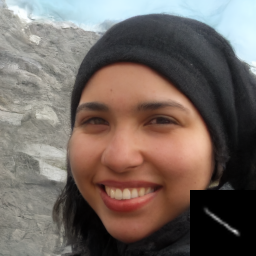} &
    \includegraphics[width=0.125\linewidth]{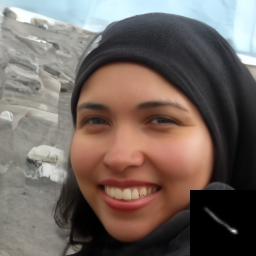} &
    \includegraphics[width=0.125\linewidth]{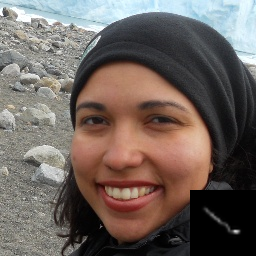} \\
        \includegraphics[width=0.125\linewidth]{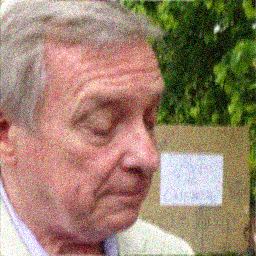} &
    \includegraphics[width=0.125\linewidth]{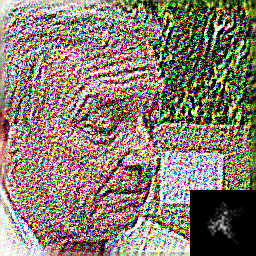} &
    \includegraphics[width=0.125\linewidth]{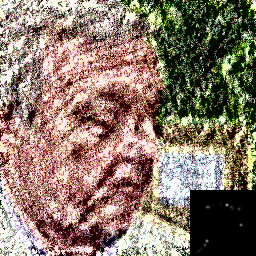} &
    \includegraphics[width=0.125\linewidth]{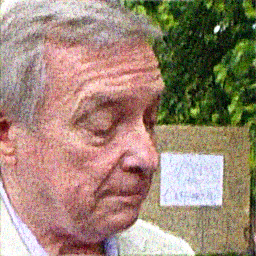} &
    \includegraphics[width=0.125\linewidth]{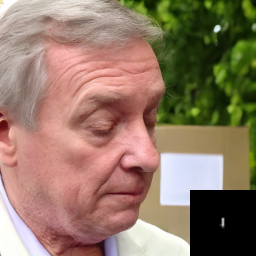} &
    \includegraphics[width=0.125\linewidth]{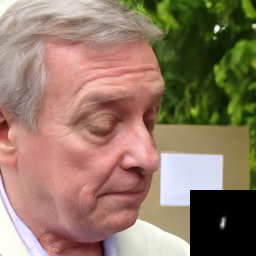} &
    \includegraphics[width=0.125\linewidth]{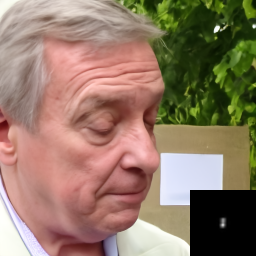} &
    \includegraphics[width=0.125\linewidth]{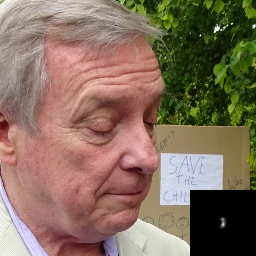} \\
        \includegraphics[width=0.125\linewidth]{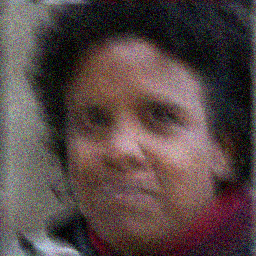} &
    \includegraphics[width=0.125\linewidth]{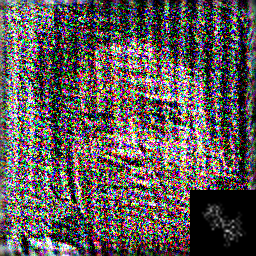} &
    \includegraphics[width=0.125\linewidth]{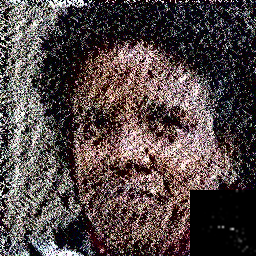} &
    \includegraphics[width=0.125\linewidth]{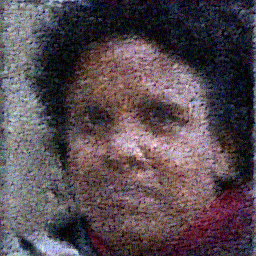} &
    \includegraphics[width=0.125\linewidth]{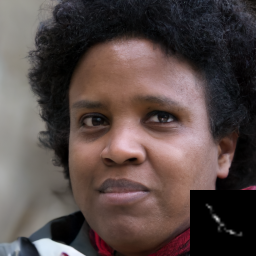} &
    \includegraphics[width=0.125\linewidth]{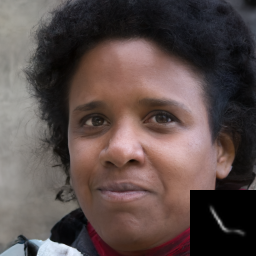} &
    \includegraphics[width=0.125\linewidth]{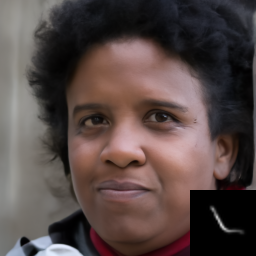} &
    \includegraphics[width=0.125\linewidth]{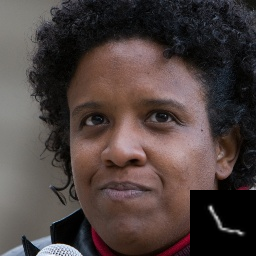} \\
        \includegraphics[width=0.125\linewidth]{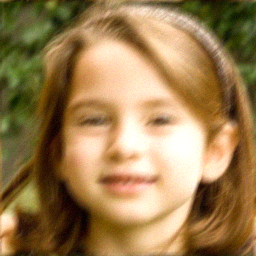} &
    \includegraphics[width=0.125\linewidth]{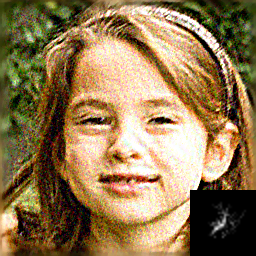} &
    \includegraphics[width=0.125\linewidth]{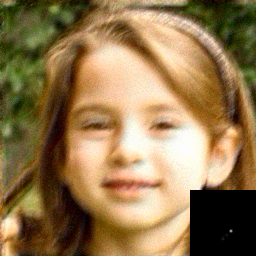} &
    \includegraphics[width=0.125\linewidth]{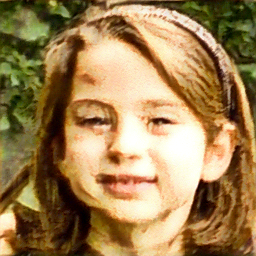} &
    \includegraphics[width=0.125\linewidth]{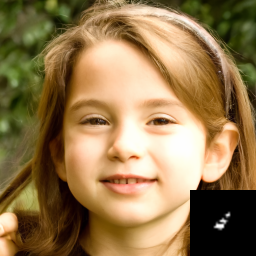} &
    \includegraphics[width=0.125\linewidth]{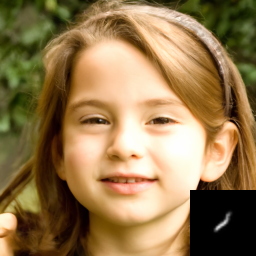} &
    \includegraphics[width=0.125\linewidth]{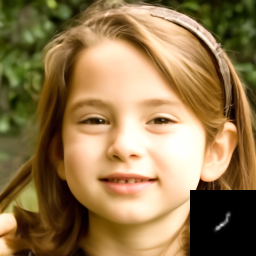} &
    \includegraphics[width=0.125\linewidth]{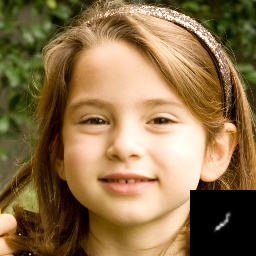} \\
        \includegraphics[width=0.125\linewidth]{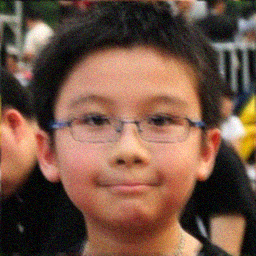} &
    \includegraphics[width=0.125\linewidth]{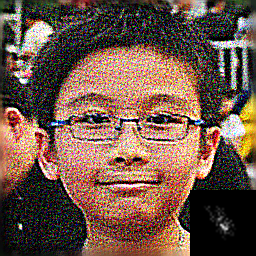} &
    \includegraphics[width=0.125\linewidth]{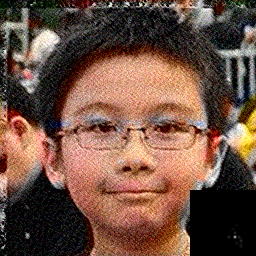} &
    \includegraphics[width=0.125\linewidth]{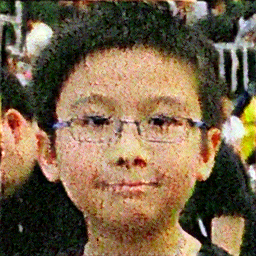} &
    \includegraphics[width=0.125\linewidth]{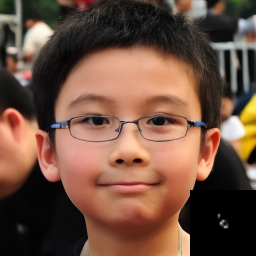} &
    \includegraphics[width=0.125\linewidth]{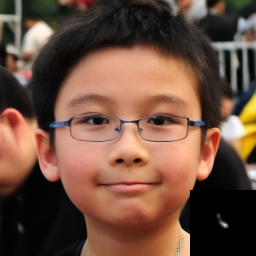} &
    \includegraphics[width=0.125\linewidth]{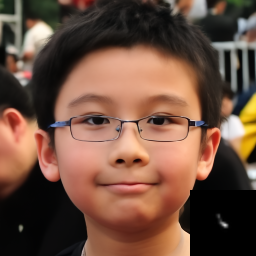} &
    \includegraphics[width=0.125\linewidth]{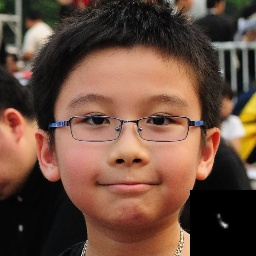} \\
        \includegraphics[width=0.125\linewidth]{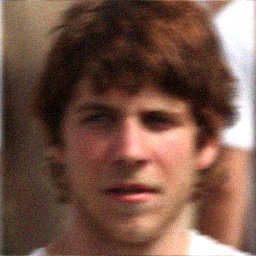} &
    \includegraphics[width=0.125\linewidth]{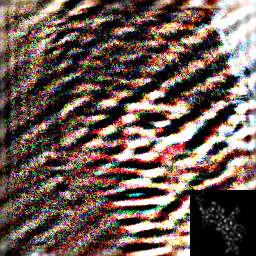} &
    \includegraphics[width=0.125\linewidth]{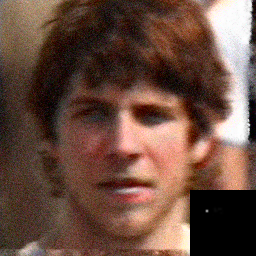} &
    \includegraphics[width=0.125\linewidth]{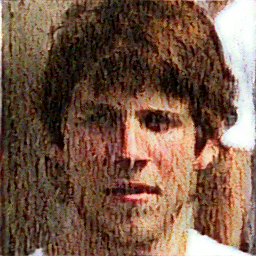} &
    \includegraphics[width=0.125\linewidth]{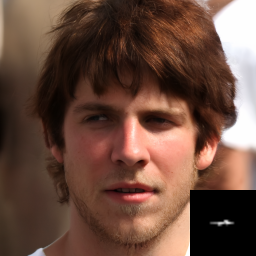} &
    \includegraphics[width=0.125\linewidth]{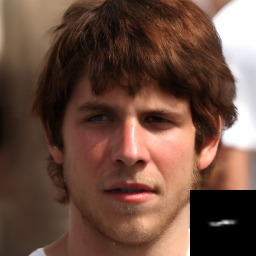} &
    \includegraphics[width=0.125\linewidth]{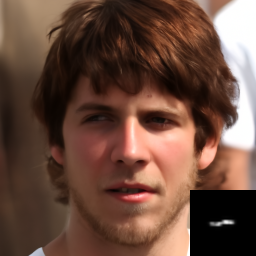} &
    \includegraphics[width=0.125\linewidth]{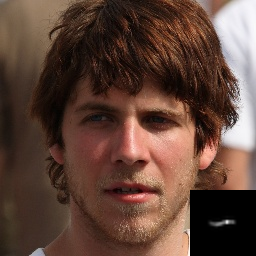} \\
        \includegraphics[width=0.125\linewidth]{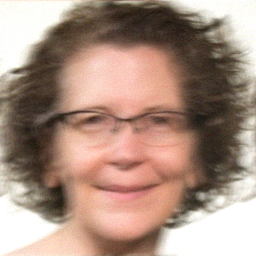} &
    \includegraphics[width=0.125\linewidth]{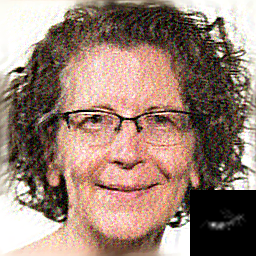} &
    \includegraphics[width=0.125\linewidth]{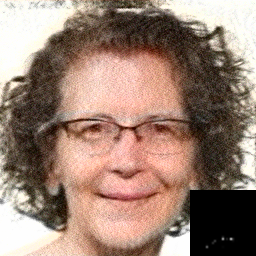} &
    \includegraphics[width=0.125\linewidth]{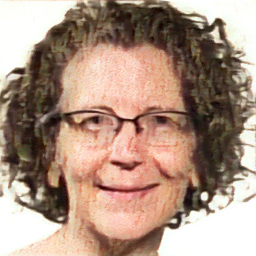} &
    \includegraphics[width=0.125\linewidth]{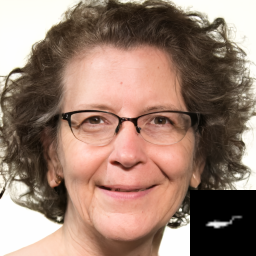} &
    \includegraphics[width=0.125\linewidth]{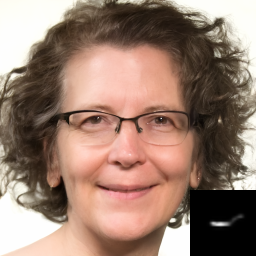} &
    \includegraphics[width=0.125\linewidth]{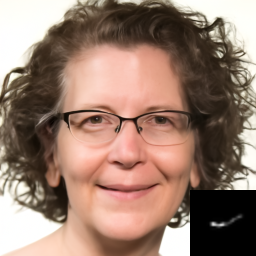} &
    \includegraphics[width=0.125\linewidth]{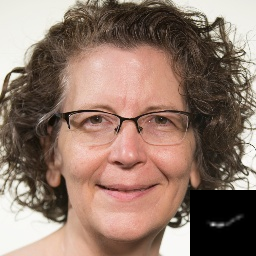} \\
\end{tabular}%
}
\caption{Visual comparison of the different models on a degraded version of FFHQ 256x256 dataset. Ours correspond to Fast EM.}
\vspace{-10pt}
\label{fig:add_res_1}
\end{figure*}
}{}

\end{document}